\pdfoutput=1

\documentclass[11pt]{article}

\usepackage[]{acl}

\usepackage{times}
\usepackage{latexsym}

\usepackage{enumitem} 
\usepackage{xcolor} 
\usepackage{arydshln}
\usepackage{geometry}
\usepackage{times}
\usepackage{amssymb}
\usepackage{lipsum}
\usepackage{latexsym}
\usepackage{adjustbox}
\usepackage{pgfplots}
\usepackage{filecontents}
\usepackage{lscape} 
\usepackage{babel}
\usepackage{rotating}
\usepackage{multirow}
\usepackage{csquotes}
\usepackage{linguex}
\usepackage{mathtools}

\usepackage[T1]{fontenc}

\usepackage[utf8]{inputenc}

\usepackage{microtype}

\title{ALEXSIS-PT: A New Resource for Portuguese Lexical Simplification}

\author{Kai North\textsuperscript{1}, Marcos Zampieri\textsuperscript{1}, Tharindu Ranasinghe\textsuperscript{2}\\
  \textsuperscript{1}George Mason University, USA \\
  \textsuperscript{2}University of Wolverhampton, UK \\
  \texttt{knorth8@gmu.edu} \\
  }

\begin{document}
\maketitle
\begin{abstract} 
Lexical simplification (LS) is the task of automatically replacing complex words for easier ones making texts more accessible to various target populations (e.g. individuals with low literacy, individuals with learning disabilities, second language learners). To train and test models, LS systems usually require corpora that feature complex words in context along with their candidate substitutions. To continue improving the performance of LS systems we introduce ALEXSIS-PT, a novel multi-candidate dataset for Brazilian Portuguese LS containing 9,605 candidate substitutions for 387 complex words. ALEXSIS-PT has been compiled following the ALEXSIS protocol for Spanish opening exciting new avenues for cross-lingual models. ALEXSIS-PT is the first LS multi-candidate dataset that contains Brazilian newspaper articles. We evaluated four models for substitute generation on this dataset, namely mDistilBERT, mBERT, XLM-R, and BERTimbau. BERTimbau achieved the highest performance across all evaluation metrics.

\end{abstract}

\section{Introduction}




The development of lexical simplification (LS) systems provides a cost-effective means of making texts accessible to individuals with reading disabilities or low-literacy who are at an economic and social disadvantage \cite{Nogueira_etal_2022}. LS systems aim to replace difficult to understand (complex) words or phrases with simpler alternatives \cite{LCPsurvey}. Consider the examples in Table \ref{example_instance}. An LS system would firstly identify the word shown in bold as being complex \cite{paetzold-specia:2016:SemEval1}. It would then generate top-k number of candidate substitutions that preserve the meaning of the original complex word in the provided context, yet are easier to understand or are more familiar to the user. These top-k candidate substitutions are then filtered based on their appropriateness, referred to as substitute generation, and finally ranked in accordance to their suitability, known as substitute ranking. 





 








Various studies have been published on Brazilian Portuguese (pt-BR) LS \cite{Aluisio2010, leal-etal-2018-nontrivial, deLima_Nascimento_2021, Leal_etal_2022}. Brazil is a country with important literacy challenges where only 1\% of its population working in the agricultural sector are proficient readers \cite{leal-etal-2018-nontrivial}. These educational challenges motivate the development of technology to assist readers and the creation of resources for pt-BR. One of the most popular pt-BR LS datasets is SIMPLEX-PB 2.0 \cite{hartmann-etal-2020-simplex_a, hartmann-etal-2020-simplex_b} featuring substitutions for over 1,500 complex words. However, in SIMPLEX-PB 2.0 only 5 ranked candidate substitutions are available for 730 of its complex words making the dataset less useful as a benchmark for state-of-the-art large pre-trained language models. Moreover, it contains texts from children's books which makes it domain-specific and therefore not a great fit for general LS. 



To address these gaps we introduce ALEXSIS-PT, a pt-BR LS dataset featuring excerpts of newspaper articles containing a larger number of candidate substitutions per target word (up to 25). To the best of our knowledge, ALEXSIS-PT has the highest average number of pt-BR ranked candidate substitutions per complex word. ALEXSIS-PT has been compiled according to the ALEXSIS protocol for Spanish \cite{Ferres_Saggion2022}, henceforth ALEXSIS-ES, opening the possibility of using cross-lingual models for these languages. ALEXSIS-PT is one of the official datasets of the TSAR shared task \cite{tsar2022}.

\begin{table*}[!ht]
\centering
\scalebox{0.85}{\begin{tabular}{l|c|c|c}
    \hline
& \textbf{SIMPLEX-PB 3.0}  & \textbf{ALEXSIS-PT} & \textbf{ALEXSIS-ES}  \\
      \hline
     Source & children's books & newspapers & newspapers  \\
    Complex words & 730 & 387 & 381 \\
    Unique complex words & 730 & 348 & 356 \\
    Annotators & 5 & 25 & 25  \\
    \textbf{Total candidate substitutions} & \textbf{3,650} & \textbf{9,605} & \textbf{9,524}  \\
    Avg. \# of total substitutions per complex word & 5 & 22 & 23 \\

     \hline
\end{tabular}}
 \caption{Comparison of ranked candidate substitutions in ALEXSIS-PT, SIMPLEX-PB 3.0, and ALEXSIS-ES.}\label{dataset_comparison}
\end{table*}

The main contributions of this paper are:
 \vspace{-2mm}
\begin{enumerate}
    \item ALEXSIS-PT, the first multi-candidate dataset for the development and evaluation of LS systems for pt-BR newspaper articles.
    \vspace{-2mm}
    \item An evaluation of multiple state-of-the-art models for LS substitute generation (SG).
\end{enumerate}

\section{Related Work}


\paragraph{Datasets and Models} The English Simple Wikipedia is an important resource that served as training material for a number of LS systems. Examples include \citet{Yatskar2010} who used Simple Wikipedia's edit history to train an unsupervised model to identify candidate substitutions for complex words and \citet{Biran2011} who trained unsupervised models on a parallel corpus with texts from Wikipedia and Simple Wikipedia texts likewise for substitute generation. Other data sources have been explored such as the Newsela corpus used in \citet{paetzold-specia-2017-lexical} who relied on neural networks together with a retrofitted context-aware word embeddings model to learn candidate substitutions. In terms of architectures, more recent LS systems use transformer-based models. \citet{qiang2020BERTLS} trained a BERT-based model to generate top-k candidate substitutions for their English dataset using masked language modelling (MLM). Others have used various transformer-based models for substitute ranking as well as precursor tasks, such as complex word identification or lexical complexity prediction \cite{LCPsurvey}.




\paragraph{Portuguese LS} The PorSimples project \cite{Aluisio2008a, Aluisio2010} sought to make online news articles more accessible in Brazil. It created the first well-known dataset for pt-BR text simplification (TS). The dataset contains excerpts of texts from a Brazilian newspaper and it is divided into 9 sub-corpora separated on degree of simplification and source text. However, unlike SIMPLEX-PB 2.0, this dataset only contained full simplified sentences and did not contain candidate substitutions for complex words needed for LS. The PorSimplesSent dataset \cite{leal-etal-2018-nontrivial} was developed to train readability classifiers to automatically predict the level of readability (complexity) of a given pt-BR sentence. This dataset was adapted from the previous PorSimples dataset \cite{Aluisio2010} but instead of presenting 9 sub-corpora with differing degrees of simplification, it combines each sentence-level simplification into pair and triple instances corresponding to original plus one or two simplifications (strong or natural). 
Finally, SIMPLEX-PB 3.0 \cite{Hartmann2020} is an extension of the previous SIMPLEX-PB 2.0 dataset \cite{hartmann-etal-2020-simplex_a, hartmann-etal-2020-simplex_b}. It added a selection of feature representations to the candidate substitutions of each complex word within SIMPLEX-PB 2.0. 

\section{ALEXSIS-PT}

\begin{table*}[!ht]
\centering
\scalebox{0.825}{\begin{tabular}{l|l}
    \hline
        \textbf{Context}& \textbf{Suggestions} \\
      \hline
    Os {\bf sedimentos} são arrastados para a parte baixa do rio. & resíduos [waste] (9), detritos [debris] (6),  lixos [garbage] (2), ...  \\
    EN: {\bf Sediments} are carried to the lower part of the river. &  fragmentos [fragments] (2), camadas [layers] (1), ...  \\
    \hline
    Simpatizantes foram {\bf arregimentados}.  & agrupados [grouped] (10), reunidos [gathered] (7), ...\\
    EN: Supporters were {\bf enlisted}. &  convocados [summoned] (2), arrebanhados [herded] (1), ... \\
    \hline
    Neste ano ocorrerão quatro ações \textbf{simultâneas} & conjuntas [joint] (7), ao mesmo tempo [at the same time] (7), ...\\
    EN: This year four {\bf simultaneous} actions will occur. &  juntas [together] (4), paralelas [parallel] (3), ...\\
    \hline
    Os partidos estão mais \textbf{cautelosos} &  cuidadosos [careful] (12), prudentes [prudent] (3), ... \\
    EN: The parties are more {\bf cautious} and careful.  &   precavidos [cautious] (3), comedidos [restrained] (1), ... \\
    \hline
    As testemunhas \textbf{contrariam} esta versão &  negam [deny] (15), desmentem [deny] (2), ... \\
    EN: The witnesses {\bf contradict} this version. & discordam [disagree] (2), desdizem [unsay] (1), ... \\

    \hline

     \hline
\end{tabular}}
 \caption{Example instances from ALEXSIS-PT. Complex words in bold, translations shown in [...], and suggestion frequency provided in (...).}\label{example_instance}
\end{table*}

We created a new dataset for pt-BR LS containing newspaper articles, referred to as ALEXSIS-PT. We did this since previous pt-BR TS datasets are either not of the newspaper genre (SIMPLEX-PB 2.0 and 3.0) or do not contain pre-identified candidate substitutions for LS (PortSimple or PorSimplesSent). ALEXSIS-PT contains a total of 387 instances with 348 of these instances having unique complex words. 
Each instance is taken from the PorSimplesSent dataset \cite{leal-etal-2018-nontrivial} and is retrieved from newspapers. PorSimplesSent is essentially a collection of original and simplified sentences thus not containing individual complex word annotations. Therefore, we had to carry out word alignment between the original and complex instances to identify complex words. This alignment was manually checked by a linguist and only the instances containing complex words that were deemed to be correctly identified were later included in the crowdsourcing platform, MTurk, for annotators to provide candidate substitutions. 

As show in Table \ref{dataset_comparison}, the choice of a relatively low number of instances but a large number of candidates (25) follows the ALEXSIS-ES protocol \cite{Ferres_Saggion2022}. The large number of total ranked candidate substitutions (9,605 against 3,650 from SIMPLEX-PB 3.0) aims to create a reliable benchmark test set for systems based on state-of-the-art large pre-trained language models. The dataset has the following format: (1) context, (2) the complex word, and (3) $n$ number of candidate substitutions (see Table \ref{example_instance}). Akin to the ALEXSIS-ES dataset, the candidate substitutions were provided by 25 Amazon MTurk annotators located in Brazil (rather than Spain) and then a careful linguistic analysis of the annotations was carried out by a linguist. In this step, 70 candidate substitutions that were either (a) equal to the complex word, (b) not pt-BR, or (c) deemed as being completely inappropriate (e.g. words that did not accurately preserve the meaning of the sentence or the original complex word) were excluded. This resulted in a final total of 9,605 candidate substitutions for the 387 complex words. These steps were also carried out by \citet{Ferres_Saggion2022} in their creation of ALEXSIS-ES.

\section{Substitute Generation}



We developed and evaluated four transformer models for substitute generation, the first step in LS pipelines. The four models are available at Hugging Face. Three of them are multilingual, being multilingual mDistilBERT (mDistilBERT) \cite{Sanh2019DistilBERTAD}, multilingual BERT (mBERT) \cite{devlin2019bert} and XLM-R \cite{conneau-etal-2020-unsupervised}, while one model is pre-trained solely on pt-BR, BERTimbau \cite{souza2020bertimbau}. We followed a similar MLM strategy to \citet{qiang2020BERTLS} where we masked the complex word of the original sentence and fed both the original sentence and the masked sentence separated by a \textsc{[SEP]} token to predict the masked token. The parameters of these models are displayed in Table \ref{models}.



We evaluated each models' ability at predicting the candidate substitutions provided by ALEXSIS-PT. Our models' were set to produce varying numbers of candidate substitutions (k) = $[1, 3, 10, 50]$ and $[1...100]$. Performance was evaluated in terms of potential, precision, recall, and F1-score where potential is the ratio of predicted candidate substitutions for which at least one of the candidate substitutions generated was among the ground truth labels. 
These evaluation metrics were chosen as they allowed for a comparison with \citet{Ferres_Saggion2022}. \citet{vstajnerlexical} has since conducted their own comparison between ALEXSIS-PT and ALEXSIS-ES as well as a third English dataset for LS. Their model's performance on these three datasets is described in Section \ref{results_here}.

The appropriateness of each models' top-k=1 candidate substitution was also evaluated by obtaining each models' average weighted frequency rank (AWFR) across all instances. AWFR shows how appropriate each top-k=1 candidate substitution is by evaluating whether it is among the top ground truth labels in terms of frequency.
We calculate AWFR as follows:

\begin{equation}
  AWFR = \frac{\sum_{n=i}^{n} f_{i} + ... + f_{n}}{\sum_{n=1}^{n} f_{1} + ... + f_{n}}
\end{equation}

\noindent where $i$ is the index of the matching ground truth label and $f$ is each ground truth labels'  corresponding frequency.

\section{Results and Discussion}\label{results_here}

\begin{table*}[!ht]
\centering
\scalebox{0.8}{\begin{tabular}{c|c|c|c|c}
    \hline
        & \textbf{mDistilBERT} &\textbf{mBERT} & \textbf{XLM\-R}  & \textbf{BERTimbau} \\
      \hline
     type & BERT-base& BERT-base & RoBERTa-base & BERT-base \\
     corpus & Wikipedia & Wikipedia & CC data &  BWaC \\
     size & 30522 Tokens & 3.3B (102 lang.) & 2.5TB (100 lang.) & 2.7B (pt-BR)\\
     \#layers & 6  & 12 & 12 & 12  \\
     \#heads & 12 & 12 & 16 & 12  \\
     \#lay.size  & 768 & 768 & 768 & 768 \\
     \#para  & 66M & 110M & 250M & 110M \\
     \hline
\end{tabular}}
 \caption{Comparison of mDistilBERT, mBERT, XLM-R, and BERTimbau models. Lang is short for languages. CC data refers to CommonCrawl data, whereas BWaC refers to the Brazilian Web as Corpus. 
 }\label{models}
\end{table*}

\begin{table*}[!ht]
\centering
\scalebox{0.8}{\begin{tabular}{c|cccc|cccc|cccc}

\hline

\multicolumn{1}{c}{\textbf{}} & \multicolumn{4}{c}{\textbf{SIMPLEX-PB 3.0}} & \multicolumn{4}{c}{\textbf{ALEXSIS-PT}} & \multicolumn{4}{c}{\textbf{Lemmatized}}\\
\hline
      \textbf{Model} & Potential & Prec.  & Recall  & F1 & Potential & Prec.  & Recall  & F1  & Potential & Prec.  & Recall & F1  \\
      
      \hline
        \multicolumn{1}{c|}{} & \multicolumn{12}{c}{top-k=1} \\
      \hline
      mDistilBERT & \textcolor{black}{0.029} & \textcolor{black}{0.029} & \textcolor{black}{0.029} & \textcolor{black}{0.029} &  0.028 & 0.028 & 0.028 & 0.028 & 0.045 & 0.045 & 0.045 & 0.045   \\

      mBERT & \textcolor{black}{0.045} & \textcolor{black}{0.045} & \textcolor{black}{0.045} & \textcolor{black}{0.045} &  0.056 & 0.056 & 0.056 & 0.056 &  0.045 & 0.045 & 0.045 & 0.045   \\
      XLM-R & \textcolor{black}{0.058} & \textcolor{black}{0.058} & \textcolor{black}{0.058}  & \textcolor{black}{0.058}  & 0.069 & 0.069  & 0.069  & 0.069 & 0.069 & 0.069  & 0.069  & 0.069  \\
      BERTimbau & \textcolor{black}{0.104} & \textcolor{black}{0.104}  & \textcolor{black}{0.104}  & \textcolor{black}{0.104} & 0.126 & 0.126  & 0.126  & 0.126  & 0.126 & 0.126  & 0.126  & 0.126 \\

      \hline
        \multicolumn{1}{c|}{} & \multicolumn{12}{c}{top-k=3} \\
      \hline

      mDistilBERT & \textcolor{black}{0.120} & \textcolor{black}{0.041} & \textcolor{black}{0.045} & \textcolor{black}{0.043} & 0.101 & 0.035 & 0.035 &  0.035 & 0.159 & 0.060 & 0.060 & 0.060   \\

      mBERT & \textcolor{black}{0.152} & \textcolor{black}{0.055} & \textcolor{black}{0.059} & \textcolor{black}{0.057} & 0.183 & 0.065 & 0.066 & 0.065  &  0.227 & 0.090 & 0.090 & 0.090   \\
      XLM-R & \textcolor{black}{0.205}  & \textcolor{black}{0.074}  & \textcolor{black}{0.080}  & \textcolor{black}{0.077}  & 0.295  & 0.112  & 0.112  & 0.112  & 0.295  & 0.113  & 0.114  & 0.114  \\
     \textbf{ BERTimbau} &  \textbf{\textcolor{black}{0.330}}  & \textbf{\textcolor{black}{0.121}}  & \textbf{\textcolor{black}{0.131}}  & \textbf{\textcolor{black}{0.126}} & \textbf{0.536}  & \textbf{0.212}  & \textbf{0.213}  & \textbf{0.213} & \textbf{0.541}  & \textbf{0.215}  & \textbf{0.216}  & \textbf{0.215} \\
      
      \hline
        \multicolumn{1}{c|}{} & \multicolumn{12}{c}{top-k=10} \\
      \hline

       mDistilBERT & \textcolor{black}{0.196} & \textcolor{black}{0.022} & \textcolor{black}{0.060} & \textcolor{black}{0.033} &  0.264 & 0.033 & 0.044 & 0.038 & 0.318 & 0.047 & 0.061 & 0.053   \\

      mBERT & \textcolor{black}{0.239} & \textcolor{black}{0.028} & \textcolor{black}{0.073} & \textcolor{black}{0.040} & 0.370 & 0.050 & 0.066 & 0.057 & 0.386 & 0.052 & 0.067 & 0.058  \\
      XLM-R & \textcolor{black}{0.316} & \textcolor{black}{0.040} &  \textcolor{black}{0.105} & \textcolor{black}{0.058} & 0.549 & 0.093 & 0.123 &  0.106  & 0.564 & 0.095 &  0.125 & 0.108 \\
      BERTimbau & \textcolor{black}{0.476} & \textcolor{black}{0.069} & \textcolor{black}{0.184} & \textcolor{black}{0.101} & 0.831 & 0.169 & 0.223 & 0.192 & 0.831 & 0.169 & 0.224 & 0.193  \\  

      \hline
        \multicolumn{1}{c|}{} & \multicolumn{12}{c}{top-k=50} \\
      \hline
       mDistilBERT & \textcolor{black}{0.266} & \textcolor{black}{0.007} & \textcolor{black}{0.099} & \textcolor{black}{0.014} &  0.422 & 0.013 & 0.089 &  0.023 & 0.431 & 0.015 & 0.102 & 0.027   \\

      mBERT & \textcolor{black}{0.299} & \textcolor{black}{0.008} & \textcolor{black}{0.109} & \textcolor{black}{0.015} & 0.512 & 0.018 &  0.120 & 0.031  & 0.500 & 0.020 & 0.129 & 0.034 \\
      XLM-R & \textcolor{black}{0.387} & \textcolor{black}{0.012} & \textcolor{black}{0.148} & \textcolor{black}{0.021} & 0.673 & 0.030 & 0.198 & 0.052 &  0.678 & 0.030 & 0.200 & 0.052 \\
      BERTimbau & \textcolor{black}{0.545} & \textcolor{black}{0.018} & \textcolor{black}{0.237} & \textcolor{black}{0.033} & 0.888 & 0.052 & 0.346 & 0.091  & 0.888 & 0.052 & 0.346 & 0.091 \\ 

     \hline
\end{tabular}}
 \caption{Substitute generation performances on the SIMPLEX-PB 3.0 and ALEXSIS-PT dataset from top-k=1 to top-k=50 candidate substitutions. Best performances are in bold.}\label{resultsSG_2}
\end{table*}

\paragraph{SG Performance} BERTimbau generated the most appropriate candidate substitutions for replacing a pt-BR complex word in any given instance. When set to generate top-k = $[1, 3, 10, 50]$ candidate substitutions, BERTimbau outperformed our mDistilBERT, mBERT and XLM-R models on all of our evaluation metrics. This is due to BERTimbau being pretrained on a single large pt-BR dataset rather than on multiple languages like mDistilBERT, mBERT and XLM-R which were found to produce candidate substitutions that were either in European Portuguese or another language entirely.

\begin{figure}[!ht]
\centering
  \includegraphics[width=0.98\linewidth]{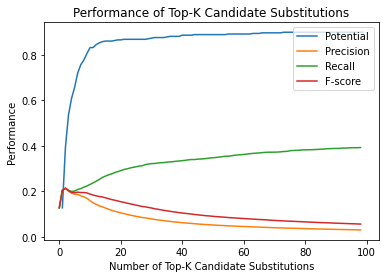}
  \caption{A plot of the BERTimbau model's potential, precision, recall, and F1-score from top-k=1 to top-k=100 candidate substitutions.}
  \label{BERTimbau_performance}
\end{figure}

Generating top-k = 3 candidate substitutions resulted in all of our models producing the highest ratio of appropriate to non-appropriate candidate substitutions. The BERTimbau model achieved a precision of 0.212 when tasked with supplying top-k=3, yet attained an inferior precision when set to return top-k= 1, 10, or 50 (Figure \ref{BERTimbau_performance}). This showed that our models were successful at predicting ground truth labels when producing a small number of candidate substitutions.

As we increase the number of top-k candidate substitutions generated we saw an increase in all models' potential and recall scores (Figure \ref{BERTimbau_performance}). Unsurprisingly, this indicates that with a greater pool of candidate substitutions, it was more likely that our models' would successfully predict multiple ground truth labels. Our BERTimbau model achieved the highest potential and recall scores of 0.888 and 0.346 respectively (Table \ref{resultsSG_2}). 

Comparing models performances across datasets, we can see that all models achieved better performances on ALEXSIS-PT in comparison to SIMPLEX-PB 3.0, with the exception of mDistilBERT when set to generate top-k = $[1, 3]$ candidate substitutions. There are two likely explanations: (1). SIMPLEX-PB 3.0 is domain specific and therefore pretrained language models may be unable to simplify vocabulary or jargon related to its genre of children's texts, and/or (2). SIMPLEX-PB 3.0 contains only 5 ranked candidate substitutions, thus the prediction of 10 or more candidate substitutions is less rewarding in regards to improving overall performance. 

\paragraph{ALEXSIS-ES Performance} A recent study by \citet{vstajnerlexical} compares the performances of LSBert \cite{qiang2020BERTLS} on ALEXSIS-PT, ALEXSIS-ES \cite{Ferres_Saggion2022}, as well as a third English dataset akin to the other two datasets. All three datasets consist of a similar number of instances (complex words in context) being 386 instances, and candidate substitutions per complex word. It was found that LSBert achieved the greatest accuracy of 30.8 on the English dataset, with ALEXSIS-PT achieving the second greatest accuracy of 15.5. In comparison, ALEXSIS-ES produced an inferior accuracy of 9.7.


\paragraph{Lemmatization}

To minimize the impact of pt-BR's fairly rich morphology and inflectional system in the evaluation, we reduce each candidate substitution and ground truth label to their lemmas. 
We use a Portuguese lemmatizer from SpaCy trained on the Universal Dependencies (UD) Portuguese treebank \cite{depling-2017}. Our models' performances increased across all evaluation metrics when taking lemmatized words. BERTimbau's performance increase was 0.002 F1-score when set to produce top-k=3 candidate substitutions. 
These results suggest that pt-BR SG systems benefit from lemmatization prior to substitute selection. Derivational or inflectional morpheme(s) can  be added further down-stream aiming to produce appropriate lexical simplification given a particular context.


\begin{table}[!ht]
\centering
\scalebox{0.85}{\begin{tabular}{c|c|c}
\hline
\multicolumn{1}{c}{\textbf{}} & \multicolumn{2}{c}{\textbf{AWFR}}\\
\hline
      Model & Original & Lemmatized  \\
      \hline
      mDistilBERT & 0.037 & 0.076  \\
      mBERT & 0.090 & 0.106  \\
      XLM-R & 0.114 & 0.115  \\
      \textbf{BERTimbau} & \textbf{0.183} & \textbf{0.185}  \\
 \hline

\end{tabular}}
 \caption{The average weighted frequency rank (AWFR) of the top-k=1 candidate substitution generated by each model.}\label{resultsSG_3}
\end{table}


\vspace{-2mm}
\paragraph{AWFR}

As shown in Table \ref{resultsSG_3}, the BERTimbau model achieves the highest AWFR across all instances after lemmatization, 0.185. This indicates that the order that our BERTimbau predicts its substitutions is the most alike to the frequency of the suggestions provided by the annotators. This is likely due to BERTimbau being trained on pt-BR data rather than multiple languages.




\section{Conclusion and Future Work}

This paper introduces ALEXSIS-PT. The dataset fills two important gaps in current LS literature: (1) it serves as a general benchmark dataset for pt-BR LS as it contains newspaper articles and (2) it provides a large number of ranked candidate substitutions making it well-suited to evaluate state-of-the-art large pre-trained language models. ALEXSIS-PT is currently the largest ranked multi-candidate pt-BR LS dataset that is accessible to the research community, consisting of 9,605 candidate substitutions.

We tested multiple models on the dataset and we report that BERTimbau achieved the best performance at SG on this new dataset. We hypothesize that this is because BERTimbau is trained only on pt-BR data while the other models were trained using multilingual data containing multiple varieties of Portuguese. Models also achieved greater performances on our new dataset in comparison to SIMPLEX-PB 3.0. We believed this to be a consequence of SIMPLEX-PB 3.0's domain specificity and its small number of ranked candidate substitutions. Lastly, we evaluated the impact of morphology in SG. Our results suggest that future SG systems developed for pt-BR should lemmatize  their output prior to substitute selection and ranking.

We are in the process of implementing a full LS pipeline on the ALEXSIS-PT dataset, including substitute selection and ranking. We also plan to explore transfer learning and develop multilingual LS systems upon the release of ALEXSIS-ES.

\section*{Acknowledgements}

We would like to thank the anonymous COLING reviewers and Matthew Shardlow for their insightful feedback. We further thank Daniel Ferr{\'e}s and Horacio Saggion, the creators of ALEXSIS, for all the information and resources they shared.

\bibliography{CWI}
\bibliographystyle{acl_natbib}

\newpage


\end{document}